\documentclass[conference]{IEEEtran}
\usepackage{graphicx}
\usepackage{tabu}
\pdfoutput=1
\usepackage{cuted, ragged2e}
\usepackage{marvosym}

\usepackage{tcolorbox}
\tcbuselibrary{most}

\usepackage{array}
\usepackage{hyperref}
\usepackage{stfloats}
\usepackage{amsmath,amssymb,amsfonts}

\usepackage{textcomp}
\usepackage{xcolor}
\usepackage{booktabs}
\usepackage{multirow}
\usepackage{xspace}
\usepackage{framed}
\usepackage{tablefootnote}
\usepackage[ruled,linesnumbered]{algorithm2e}
\usepackage{subcaption}
\usepackage{bbm}
\usepackage{xspace}
\usepackage{balance}

\usepackage{marginnote}
\newcommand{\marginX}{\marginnote{\huge{\quad\quad\textbf{!}\quad\quad}}}
\newcommand{\youcheng}[1]{{\color{blue}\marginX{}[\textbf{Youcheng}: #1]}}
\usepackage{booktabs}
\newcommand{\lab}[1]{\textquotesingle{#1}\textquotesingle}

\usepackage{xspace}
\newcommand{\airepair}{{\sc AIRepair}\xspace}
\newcommand{\commentout}[1]{}

\begin{document}

\title{\airepair: A Repair Platform for Neural Networks}
\author{
\IEEEauthorblockN{Xidan Song\IEEEauthorrefmark{1}, Youcheng Sun\IEEEauthorrefmark{1}, Mustafa A. Mustafa\IEEEauthorrefmark{1}\IEEEauthorrefmark{2}, Lucas C. Cordeiro\IEEEauthorrefmark{1}\IEEEauthorrefmark{3}}

\IEEEauthorblockA{\IEEEauthorrefmark{1}Department of Computer Science, University of Manchester, UK}

\IEEEauthorblockA{\IEEEauthorrefmark{2}imec-COSIC, KU Leuven, Belgium}

\IEEEauthorblockA{\IEEEauthorrefmark{3}Federal University of Amazonas, Brazil}

\IEEEauthorblockA{\IEEEauthorrefmark{1}{\{xidan.song, youcheng.sun, mustafa.mustafa, lucas.cordeiro\}@manchester.ac.uk}}
}

\maketitle

\begin{abstract}
    We present \airepair, a platform for repairing neural networks. It features the integration of existing network repair tools.
    Based on \airepair, one can run different repair methods on the same model, thus enabling the fair comparison of different repair techniques. 
    In this paper, we evaluate \airepair with five recent 
    repair methods on popular deep-learning datasets and models. Our evaluation confirms the utility of \airepair, by comparing and analyzing the results from different repair techniques. A demonstration is available at \url{https://youtu.be/UkKw5neeWhw}. 
\end{abstract}

\section{Introduction}
\label{sec:introduction}

As neural networks are widely applied in various areas, 
insufficient accuracy, adversarial attack, and data poisoning, among others, threaten the development and safe deployment of neural network applications~\cite{huang2020survey}. Unlike traditional software, a neural network model is not programmed manually by software developers. Instead, its parameters are automatically learned from training data. 
As a result, these models' defects cannot be easily fixed.
 
Many repair techniques and tools, like DeepRepair~\cite{yu2021deeprepair}, DL2~\cite{fischer2019dl2}, Apricot~\cite{zhang2019apricot}, DeepState~\cite{liu2022deepstate} and RNNRepair~\cite{xie2021rnnrepair} have been proposed to automatically fix the defects found in a neural network 
Naturally, the neural network repair has two objectives: correcting the model's wrong behaviors on failed tests while not comprising its original performance on passing tests. Existing repair techniques can be classified into three categories: retraining/fine-tuning, direct weight modification, and architecture extension. 

In the first category of repair methods, the idea is to retrain or fine-tune the model for the corrected output with the identified misclassified input. 
Typical methods include counterexample-guided data augmentation, which adds misclassified examples iteratively to the
training datasets~\cite{dreossi2018counterexample}. 
The editable training in~\cite{sinitsin2020editable} aims to efficiently patch a mistake of the model on a particular sample, and the training input selection in~\cite{ma2018mode} emphasizes selecting high-quality training samples for fixing model defects. 
DeepRepair~\cite{yu2021deeprepair} implements transfer-based data augmentation to enlarge the training dataset before fine-tuning the models. RNNRepair~\cite{xie2021rnnrepair} and DeepState~\cite{liu2022deepstate} also augment the datasets and perform retraining to improve the performance of the neural network models; they are specifically designed to repair the Reconcurrent Neural Network (RNN)~\cite{svozil1997introduction} models.

The methods in the second category calculate new weight values of the model to fix erroneous nodes in the neural network, to improve the classification accuracy, or to satisfy safety properties. In~\cite{goldberger2020minimal} and \cite{usman2021nn}, SMT solvers are used for solving the weight modification needed at the output layer for the neural network to meet specific requirements without any retraining. In~\cite{sohn2022arachne}, the repair method localizes the faulty neuron weights and optimizes the localized weights to correct the model misbehavior by Particle Swarm Optimisation algorithm. 
The Apricot tool~\cite{zhang2019apricot} adjusts the weights of a model with the feedback from a set of so-called reduced models trained with subsets of training data.

Repair techniques in the third category extend a given neural network architecture, e.g., by introducing more weight parameters or repair units, to facilitate more efficient repair. PRDNN~\cite{sotoudeh2021provable} introduces Decoupled DNNs, a new DNN architecture that enables efficient and effective repair. DeepCorrect~\cite{borkar2019deepcorrect} corrects the worst distortion-affected filter activations by appending correction units. DL2~\cite{fischer2019dl2} implements neural networks that can train with constraints that restrict inputs outside the training set.
Researchers have used these different methods to develop many repair tools. Different repair tools run on different environments with customized configurations. They accept different input formats and produce various outputs. It can be challenging 
to figure out which repair tool and setting to use. 
To make it easier for users to configure appropriate repair methods and parameters, we developed a neural network repair platform -- \airepair. It can automatically repair trained models using different methods. 
We believe that \airepair would be helpful for developers who want to improve their neural networks and those interested in evaluating their new repair methods. 

The main contributions of this paper are three-fold:
\begin{itemize}
    \item We develop \airepair for integrating and evaluating existing (and future) repair techniques on neural networks.
    \item We benchmark five repair techniques on 11 types of neural network models across four datasets.
    \item We make \airepair and its benchmark publicly available: \url{https://github.com/theyoucheng/AIRepair}
\end{itemize}


\section{\airepair Description}
\label{Description}

\subsection{Overview}

\begin{figure}
    \centering
        \centering
        \includegraphics[width=\linewidth]{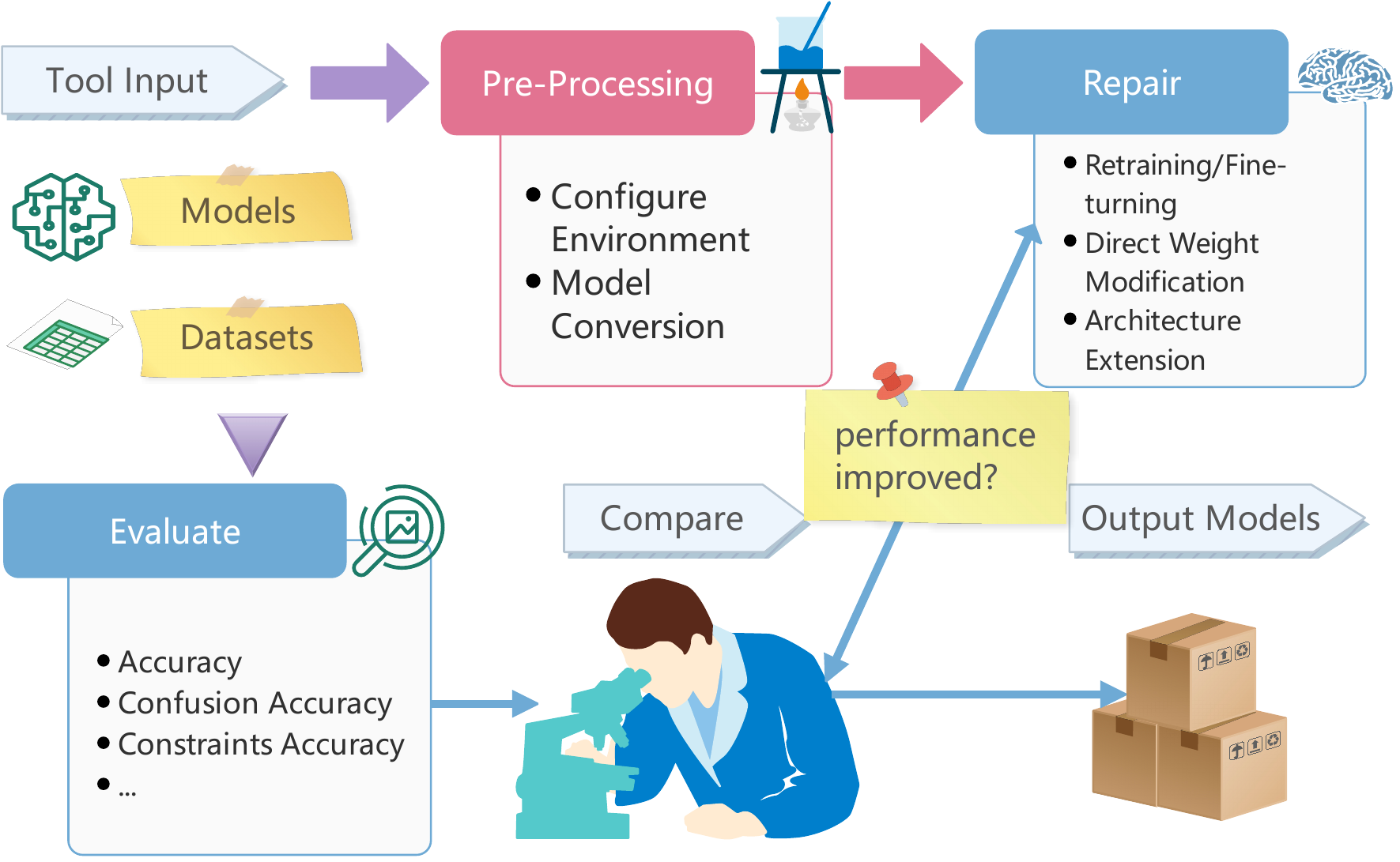}
    \caption{The \airepair Architecture.}
    \label{fig:property}
\end{figure}

The motivation for \airepair is to develop a platform for testing and evaluating different repair methods in a compatible way. 
As illustrated in Fig.~\ref{fig:property}, \airepair accepts the trained models and the training datasets if they are specified in the configuration. It performs pre-processing on different benchmarks to make them capable of different frameworks. Pre-processing isolates different running environments for various deep learning libraries, e.g., TensorFlow~\cite{abadi2016tensorflow} or PyTorch~\cite{paszke2019pytorch}. After the repair, \airepair collects the results and analyses them automatically, which is done by examining the outputs and experimental logs. Finally, it presents the results for the user to decide which repairing tool suits their models. The output from \airepair includes the repaired model combined with the logs and parameters.

\subsection{Modular View}
\label{sec:tool_modular}

In this part, we detail each component in \airepair: 1) input, 2) pre-processing, 3) repair, and 4) evaluation.

\subsubsection{Input}\label{sec:tool_modular_input} The input to our platform is trained neural network models and testing or training datasets depending on the repair method configured. \airepair accepts fully connected feed-forward neural networks and convolutional neural networks in popular deep learning model formats such as .pth and .pt from PyTorch, and .pb, and .h5 from TensorFlow + Keras~\cite{chollet2015keras}. Our platform has been tested on standard datasets like MNIST, Fashion-MNIST, CIFAR-10 and CIFAR-100. 

\subsubsection{Pre-processing}
This component converts neural network models between different formats, configures the running environments for various underlying repair tools, and evaluates the models before repair. 
\subsubsection{Repair}
Based on the input and pre-processing, the repair component configures and calls the underlying repair methods to perform repairs on the input models. 
 During repair, it monitors the hardware resource consumption, displays messages for the status of the repair procedure, and saves the logs for evaluation.

\subsubsection{Evaluation}
This component measures the performance of a model before and after repair. There are several metrics for characterizing a model's performance, from complementary perspectives, including the model's \emph{(classification) accuracy}, \emph{constraint accuracy}~\cite{fischer2019dl2}, and \emph{confusion accuracy}. The constraint accuracy describes the percentage
of predictions given by the model that satisfies the constraint associated with the problem, which requires that the probabilities of groups of classes have either a very high or a very low probability. The confusion accuracy is defined as ${P=\frac{TP}{TP+FP}}$, where $TP$ and $FP$ are True Positive and False Positive classifications. These two metrics evaluate the model's robustness. 
Data augmentation of different kinds of blurs (glass, motion, and zoom)~\cite{laugros2019adversarial} can be applied on the input dataset when collecting these metrics.


Typically, existing repair tools take some misclassified inputs, and the repair goal is to correct those erroneous nodes.
Each repair method often focuses on improving the performance according to one type of evaluation metric. The \airepair tool integrates different repair methods and different performance metrics, to give a comprehensive view when repairing a neural network model.  

\subsubsection{Output}
\airepair delivers the repaired models after repair and the resulting configurations for each repair method that will be stored in log files. 

\subsection{Example Usage}
\label{sec:tool_example}

At the first step, we encourage one to train the baseline model using the script (\lab{train\_baseline.py}) provided.
\commentout{
\begin{lstlisting}[xleftmargin=2em,frame=single,framexleftmargin=1.5em, basicstyle=\small]  
python train_baseline.py 
[-h] [--all ALL] [--tool TOOL]
[--net_arch NETARCH] 
[--dataset DATASET]
[--additional_param PARAM] 
[--saved_path PATH]
\end{lstlisting}
}
Subsequently, one can configure and run different network repair tools with \airepair:

\begin{lstlisting}[xleftmargin=2em,frame=single,framexleftmargin=1.5em, basicstyle=\small]  
python AIRepair.py [-h] [--all]
[--net_arch NETARCH] [--dataset DATASET]
[--pretrained PATH_AND_FILENAME] 
[--depth DEPTH]
[--method METHOD] [--auto] 
[--additional_param PARAM] 
[--input_logs INPUT_LOGS] 
[--testonly]
\end{lstlisting}

For example, the setup below configures and runs \airepair with three repair methods, Apricot, DeepRepair, and DL2, as discussed in Section \ref{sec:introduction}. They are applied to repair a model named \lab{cifar10\_resnet34} with the CIFAR-10 dataset.
\begin{lstlisting}[xleftmargin=2em,frame=single,framexleftmargin=1.5em, basicstyle=\small, breaklines=true]  
python AIRepair.py --method apricot deeprepair dl2 --pretrained cifar10_resnet34_baseline.pt --dataset cifar10 --net_arch resnet --depth 34
\end{lstlisting}

In particular, when using \lab{-{}-pre-trained} to specify the path  of a trained neural network model to repair, users need to specify the model's architecture as well as depth. PyTorch has two different methods to save the trained model: saving the entire model or saving the state\_dict or checkpoint. When loading the neural network model, \airepair needs to know its structure for the second method. It has the built-in structure definition for ResNet families and several convolutional neural networks for MNIST and Fashion-MNIST. Hence, users only need to specify the net architecture and depth when loading the state\_dict. For the architectures that do not belong to these three models, users either provide the entire model or customize \airepair's pre-processing module. Currently, \airepair can process both feed-forward neural networks and convolutional neural networks.

The parameter \lab{-{}-net\_arch} specifies the architecture of models, and \lab{-{}-dataset} selects the corresponding dataset (that are needed for retraining/refining or attaching correction units). These are as discussed in Section \ref{sec:tool_modular_input}.
The specific repair method (tool) can be specified using \lab{-{}-method}, and  
\lab{-{}-auto} will automatically invoke the repair using the default parameters for the selected repair methods. For example, it sets the following configurations for DeepRepair method to repair ResNet34 trained on CIFAR-10 dataset. 

\begin{lstlisting}[xleftmargin=2em,frame=single,framexleftmargin=1.5em, basicstyle=\small, breaklines=true]  
--batch_size 128 --lr 0.1 --lam 0 
--extra 128 --epoch 60 --beta 1.0 
--cutmix_prob 0 --ratio 0.9
\end{lstlisting}

Alternatively, one can use \lab{-{}--additional\_param} to customize these parameters, which will substitute these default settings.  
The \lab{-\--testonly} option is for evaluating a model without any repair procedure; users can use it before or after repair to check the model's performance. The evaluation metrics include accuracy, confusion, and constraint accuracy.
For convenience,  \lab{python AIRepair.py  -{}-all} runs all the repair methods on all available models \footnote{\url{https://zenodo.org/record/7627801##.Y-X6g3bP3tU}} with default parameters automatically. Note that this option requires substantial computing power. 
\section{Experiment}
\label{sec:experiment}










\subsection{Experimental Setup}

We ran experiments on a machine with Ubuntu 18.04.6 LTS OS Intel(R) Xeon(R) Gold 5217 CPU @ 3.00GHz and two Nvidia Quadro RTX 6000 GPUs. 
Five repair methods, Apricot~\cite{zhang2019apricot}, DeepRepair~\cite{yu2021deeprepair}, DL2~\cite{fischer2019dl2}, DeepState~\cite{liu2022deepstate}, and RNNRepair~\cite{xie2021rnnrepair}, are chosen as baselines. They are applied to repair a benchmark of 11 neural network models (including ResNet models~\cite{he2016deep}) with four datasets 
MNIST~\cite{deng2012mnist}, Fashion-MNIST~\cite{xiao2017fashion}, CIFAR-10 and CIFAR-100~\cite{krizhevsky2009learning}.
Users can invoke it by specifying \lab{-{}-auto} as discussed in Section \ref{sec:tool_example}. 
We repeat the same experiment three times to eliminate randomness in repair methods.

\begin{table*}[t]
\centering
\caption{AIRepair results: 
The best accuracy (Acc.) and constrains accuracy (Const.) improvement for each model is highlighted in \fcolorbox{blue}{white}{\color{blue}\%} and \fcolorbox{red}{white}{\color{blue}\%} separately.}
\label{table:final_results}
\begin{tabular}{ccccccccccc} 
\toprule
\multicolumn{2}{c}{Datasets}                                                                                                   & \multicolumn{1}{l}{} & \multicolumn{3}{c}{CIFAR-10}    & \multicolumn{3}{c}{CIFAR-100}   & MNIST   & Fashion-MNIST  \\ 
\midrule
\multicolumn{2}{c}{Models}                                                                                                     & \multicolumn{1}{l}{} & ResNet18 & ResNet34 & ResNet50 & ResNet18 & ResNet34 & ResNet50 & MNIST & Fashion-MNIST        \\ 
\midrule
\multicolumn{2}{c}{\multirow{2}{*}{Baselines}}                                                                                  & Acc.                 & 92.05\%  & 91.34\%  & 94.42\%  & 46.84\%  & 44.16\%  & 47.36\%  & 99.45\% & 92.20\%       \\
\multicolumn{2}{c}{}                                                                                                           & Const.               & 90.51\%  & 90.27\%  & 90.66\%  & 86.62\%  & 85.95\%  & 85.21\%  & 99.96\% & 100\%         \\ 
\midrule
\multicolumn{2}{c}{{Apricot}}                      & Acc.                 & -2.65\%  & -0.38\%  & -3.4\%   & +9.02\%  & +13.74\% & +11.15\% & +0.06\% & \fcolorbox{red}{white}{\color{blue}+0.61\%}       \\
                                                   
\multicolumn{2}{c}{\multirow{2}{*}{DeepRepair}}  & Acc.                 & \fcolorbox{red}{white}{\color{blue}+0.5\%}   & -1.27\%  & -4.14\%  & \fcolorbox{red}{white}{\color{blue}+10.91\%} & \fcolorbox{red}{white}{\color{blue}+21.42\%} & \fcolorbox{red}{white}{\color{blue}+20.32\%} & \fcolorbox{red}{white}{\color{blue}+0.17\%}  & +0.47\%       \\
                                                                                                 &                             & Const.               & -9.46\%  & -8.82\%  & -12.77\% & -37.62\% & -34.95\% & -29.71\% & -0.43\% & -4.10\%       \\
\multicolumn{2}{c}{\multirow{2}{*}{DL2}}         & Acc.                 & -2.16\%  & \fcolorbox{red}{white}{\color{blue}+0.23\%}  & -1.95\%  & \fcolorbox{blue}{white}{\color{blue}+0.87\%}   & \fcolorbox{blue}{white}{\color{blue}+1.17\%}  & -1.16\%  & +0.08\% & +0.28\%       \\
\multicolumn{1}{l}{}                                                                             &                             & Const.               & \fcolorbox{blue}{white}{\color{blue}+9.3\%}   & \fcolorbox{blue}{white}{\color{blue}+9.61\%}  & \fcolorbox{blue}{white}{\color{blue}+5.4\%}   & -0.49\%  & -0.89\%  & -0.4\%   & \fcolorbox{blue}{white}{\color{blue}+2.55\%} & \fcolorbox{blue}{white}{\color{blue}+6.27\%}           \\

\bottomrule
\end{tabular}
\end{table*}

\commentout{
\begin{table}[t]
\centering
\caption{AIRepair RNN model results}
\label{table:final_results_rnn}
\scalebox{0.8}{
\begin{tabular}{lllllll} 
\toprule
Datasets       & \multicolumn{3}{c}{MNIST}    & \multicolumn{3}{c}{Fashion-MNIST}  \\
\midrule
Models         & LSTM     & BLSTM   & GRU     & LSTM    & BLSTM & GRU \\
\midrule
Baseline\_ori  & 98.6\%   & 98.54\% & 98.93\% & 88.34\% & 89.10\%     & 88.90\%          \\
DeepState      & +0.05\%  & +0.19\% & +0.18\% & +1.12\% & -0.15\%      & +0.72\%          \\
RNNRepair      & -0.09\%  & -0.06\%        &  -0.33\%       &         &       &                  \\
Baselines\_mix & 66.95\%  &  69.67\%       &         &         &       &                  \\
DeepState      & +12.12\% &  +9.76\%       &         &         &       &                  \\
RNNRepair      & +9.79\%        & +4.25\%        &         &         &       &                  \\
\bottomrule
\end{tabular}}
\end{table}
}

\subsection{Results}
\label{sec:experiment_results}

\commentout{
\subsubsection{Train baseline models}

We used the DL2 tool to train the baseline models. We trained three Convolutional Neural Network models with different depths, ResNet18, ResNet34 and ResNet50 ~\cite{he2016deep}, on the CIFAR-10, CIFAR-100 datasets separately. We also trained two simple convolutional Neural Network Models on MNIST and Fashion-MNIST datasets. The details of these models are shown in Table \ref{table:all_benchmarks}. We set the weight of the DL2 abstraction layer to 0 to serve as the baseline network, and all subsequent neural network repairing was performed on the baseline neural network models.

\subsubsection{Compare different repairing tools on the same benchmark}
}

Table \ref{table:final_results} shows the complete \airepair results of 3 repair methods on 8 neural networks (columns) from 4 datasets. We measured the performance by accuracy and constraint accuracy. The baseline represents each model's actual performance, and for each repair method, we report the absolute performance increment or decrement after repair. Apricot repairs a neural network by directly modifying its weights parameter values. DeepRepair and DL2 belong to the category of retraining and architecture extension, respectively, for repairing neural networks. The three baselines are selected to represent all 5 categories of neural network repair methods, as discussed in Section \ref{sec:introduction}.  

For the models trained on CIFAR-10, we can see that DL2 brings the highest constrains accuracy (Const.) improvement, while this often comes with slight drops in the plain accuracy (Acc.). 
The adversarial robustness and the plain accuracy are two contradicting goals~\cite{tsipras2018there}. Besides, the performance of the CIFAR-10 models drops to different extents by Apricot and DeepRepair.
Apricot is not designed to improve the constrains accuracy of models. We do not evaluate the constrains accuracy metrics on models repaired by Apricot.

At the same time, Apricot and DeepRepair have better improvements on CIFAR-100 neural network models. This indicates that when the original model's performance is relatively low, the Apricot and DeepRepair are more effective. Moreover, as the ResNet models become deeper, from 18 layers to 50 layers, DeepRepair can improve accuracy by sacrificing constraint accuracy.
DL2 has no significant improvement in performance this time. 

The original implementations for Apricot, DeepRepair, and DL2 do not support models trained on MNIST and Fashion-MNIST, so we created patches to integrate them into \airepair for pre-processing and repairing these datasets and models.
As shown in Table \ref{table:final_results} (last two columns), all three repair methods result in noticeable performance improvements of different levels for MNIST and Fashion-MNIST models, even though the original networks' performance is already high (99.45\% and 92.20\% respectively).


In summary, there are some general observations from the experiments. There is no single best repair method on all benchmarks and on all evaluation metrics. Different repair methods seem to complement each other highly, and this would motivate future ``combined repair'' of neural network models, for which \airepair can serve as the test bed. Mostly, as expected, less complex neural networks (MNIST/Fashion-MNIST in Table \ref{table:final_results}) and lower performance models (CIFAR-100 examples in Table \ref{table:final_results}) are easier to repair. We still regard them as valuable observations, as they confirm that \airepair is a valid platform for benchmarking different repair methods.  

\paragraph*{More neural network architectures}
We also tested \airepair on three RNN-based architectures. We apply DeepState and RNNRepair to repair LSTM, BLSTM, and GRU models trained on MNIST.
The baseline 
model classification accuracy is $98.6$\%, $98.54$\%, and $98.93$\%. DeepState slightly improves the accuracy of these models by $0.05$\%, $0.19$\%, and $0.18$\%, whereas the model accuracy drops by $0.09$\%, $0.06$\%, and $0.33$\% when using RNNRepair. 

\subsection{Discussions}

Based on \airepair experiments, when encountering the performance anomaly of a neural network model, we suggest using neural network repair methods in the following order: 1) direct weight modification, 2) fine-tuning/retraining, and 3) attaching a new repair structure into the model. 

For simpler neural networks like models trained on MNIST and Fashion-MNIST datasets, we put direct weight modification as a higher priority for repair. This repair does not require training/testing datasets and can be done without GPU hardware. However, it may face more significant challenges when repairing a trained model with high performance. 

For complex convolutional models such as ResNets 
with millions of parameters, direct weight modification is more likely to encounter the search space explosion problem. In such cases, retraining/refining-based repair methods can still be applied without altering the neural network's structure. However, they may require more powerful hardware support.


If there is over-fitting during the retraining procedure or even further improvement is needed for the repair, or the models need to improve specific constraint accuracies, one could consider attaching repairing units.

Determining which repair method to use may depend on the NN application scenarios. For example, NN models trained on the MNIST dataset recognize handwriting materials, while models trained on ImageNet are designed to be used on image-based searches. We suggest referring to the above discussion and prioritizing the selection of the appropriate repair according to the characteristics and types of models. Another unique use case is the quantized neural network models used in mobile and embedded devices.

\section{Conclusion}
\label{Conclusion}

We present \airepair, a comprehensive platform for repairing neural networks, and it can test and compare different neural network repair methods. This paper gives the results of five existing neural network repair tools integrated into \airepair. Although \airepair is an early prototype, it shows promising results. We will support and test more neural network repair methods and propose a unified interface for developers to test and benchmark their repair methods. 

\section*{Acknowledgements}

This work is funded by the EPSRC grant EP/T026995/1 \textit{``EnnCore: End-to-End Conceptual Guarding of Neural Architectures''}. The Dame Kathleen Ollerenshaw Fellowship of The University of Manchester supports M. A. Mustafa.

\balance
\bibliographystyle{IEEEtran}
\bibliography{references}

\end{document}